\documentclass[11pt]{article}

\usepackage[preprint]{acl}

\usepackage{times}
\usepackage{latexsym}

\usepackage[T1]{fontenc}

\usepackage[utf8]{inputenc}

\usepackage{microtype}

\usepackage{inconsolata}
\usepackage{multirow}
\usepackage{booktabs}

\usepackage{graphicx}
\usepackage{xurl}
\usepackage{amsmath}

\title{AgentFactory: A Self-Evolving Framework Through Executable Subagent Accumulation and Reuse}

\author{
  \textbf{Zhang Zhang\textsuperscript{1,3}},
  \textbf{Shuqi Lu\textsuperscript{3}},
  \textbf{Hongjin Qian\textsuperscript{1,3}},
  \textbf{Di He\textsuperscript{1}},
  \textbf{Zheng Liu\textsuperscript{2,3}}
\\
\\
  \textsuperscript{1}Peking University,
  \textsuperscript{2}Hong Kong Polytechnic University
\\
  \textsuperscript{3}Beijing Academy of Artificial Intelligence
\\
  \texttt{zzhang@stu.pku.edu.cn}, \texttt{zhengliu1026@gmail.com}
}

\begin{document}
\maketitle
\begin{abstract}
Building LLM-based agents has become increasingly important. Recent works on LLM-based agent self-evolution primarily record successful experiences as textual prompts or reflections, which cannot reliably guarantee efficient task re-execution in complex scenarios. We propose AgentFactory, a new self-evolution paradigm that preserves successful task solutions as executable subagent code rather than textual experience. Crucially, these subagents are continuously refined based on execution feedback, becoming increasingly robust and efficient as more tasks are encountered. Saved subagents are pure Python code with standardized documentation, enabling portability across any Python-capable system. We demonstrate that AgentFactory enables continuous capability accumulation: its library of executable subagents grows and improves over time, progressively reducing the effort required for similar tasks without manual intervention. Our implementation is open-sourced at \url{https://github.com/zzatpku/AgentFactory}, and our demonstration video is available at \url{https://youtu.be/iKSsuAXJHW0}.
\end{abstract}

\section{Introduction}
\label{sec:intro}

Large Language Models (LLMs) have demonstrated remarkable capabilities in reasoning, planning, and problem-solving \cite{anthropic2026claude46, openai2026gpt54systemcard, gemini31}. The development of LLM-based agents that can interact with external tools and environments has become a critical research area \cite{yao2023reactsynergizingreasoningacting,wang2024survey,xi2025rise}. These agents must not only reason about tasks but also execute actions, retrieve information, and adapt to dynamic environments.

Existing frameworks for building LLM agents, such as LangChain \cite{langchain2022} and AutoGPT \cite{autogpt2023}, provide valuable abstractions for connecting LLMs with external tools. However, these frameworks treat agent behavior as static---knowledge gained during execution is not preserved for future use. To address this, recent works have explored self-evolving agents that record and leverage past experience through verbal reflection, iterative output refinement, or reasoning bootstrapping \cite{shinn2023reflexion,madaan2023self,zelikman2022star}. However, these approaches primarily record successful experiences as textual prompts, verbal reflections, or reasoning traces. For complex real-world tasks, such textual experience cannot reliably guarantee efficient task re-execution. Recent advances have begun exploring code-based self-evolution as an alternative. AlphaEvolve \cite{novikov2025alphaevolve} demonstrates the power of code-based evolution for high-complexity scientific discovery, and the Darwin Gödel Machine \cite{zhang2025darwin} explores open-ended recursive self-improvement of agent internals. However, these methods are primarily tailored for solving highly specialized scientific or meta-reasoning problems. Nonetheless, a vast array of general-purpose workflows still relies on routine, procedural tasks that demand a different approach to evolution.


We observe that although daily user tasks appear diverse and complex, most can be decomposed into a set of reusable subtasks---for instance, scheduling a meeting, conducting a literature survey, or file manipulations. This observation motivates us to propose a self-evolution approach centered on creating and accumulating subagents: we decompose tasks into subtasks and construct specialized subagents for each one. Successfully executed subagents are saved as executable code, forming a growing library of reusable capabilities. Crucially, these subagents are autonomously refined based on execution feedback from subsequent tasks, making them increasingly robust and general-purpose over time. Furthermore, since all saved subagents are pure Python code with standardized documentation, they can be exported and deployed across any Python-capable system, enabling cross-platform capability transfer. 

Based on this design, we propose \textbf{AgentFactory}, a self-evolving framework that implements a three-phase lifecycle:
(1) \textbf{Install}: Construct subagents from scratch to solve initial problems, building a library of reusable subagents.
(2) \textbf{Self-Evolve}: When encountering similar tasks, detect limitations in saved subagents and autonomously improve them to be more general-purpose and robust.
(3) \textbf{Deploy}: Export mature subagents as standalone Python modules for use in other AI frameworks.

We observe that after executing a small number of initial tasks, the system begins to accumulate a collection of subagents that can be reused to assist with many common real-world tasks. In this way, solving a few initial tasks provides a practical starting point: the subagents constructed during these early runs can be leveraged to support subsequent requests without additional manual setup. This automatically running pipeline creates an efficient self-evolution system that becomes increasingly helpful for everyday user needs. 

Our contributions are as follows:
 
\begin{enumerate}
    \item We propose a three-phase self-evolving pipeline (Install $\rightarrow$ Self-Evolve $\rightarrow$ Deploy) that systematically builds, improves, and exports functional agents. Our framework enables the system to accumulate and improve complex task-solving capabilities rather than merely recording procedural experiences.
     
    \item We implement AgentFactory and demonstrate on diverse tasks that the system becomes increasingly effective: saved subagents are directly reused to solve new problems (faster problem-solving), and they are iteratively improved based on execution feedback (growing capabilities).
\end{enumerate}

\section{Related Work}
\label{sec:related}

Our work relates to multi-agent systems, self-evolving mechanisms, and skill accumulation.

\paragraph{Multi-Agent Systems} Multi-agent frameworks such as AutoGen \cite{wu2024autogen}, MetaGPT \cite{hong2023metagpt}, and ChatDev \cite{qian2024chatdevcommunicativeagentssoftware} enable collaboration among specialized agents through predefined workflows. Recent research has shifted toward dynamic orchestration and topological optimization. For instance, AgentVerse \cite{chen2024agentverse} mimics human group dynamics for expert recruitment, while DyLAN \cite{liu2023dynamic} introduces an unsupervised metric for dynamic agent team optimization. Furthermore, GPTSwarm \cite{zhuge2024gptswarm} treats agents as optimizable graphs, and frameworks like CrewAI and LangGraph support role-based task execution and cyclic state management. 


\paragraph{Self-Evolution and Skill Accumulation} 
Recent advancements enable agents to continuously improve through self-evolution and persistent memory. Evolutionary approaches optimize specific agent components, including prompts \cite{wang2024promptagent, guo2024evoprompt, fernando2024promptbreeder}, reasoning strategies \cite{shinn2023reflexion, zelikman2022star}, architectures \cite{hu2024automated, zhang2025aflow, li2024autoflowautomatedworkflowgeneration}, and algorithmic code \cite{novikov2025alphaevolve, zhang2025darwin}. Concurrently, skill accumulation methods maintain experience through structured memory mechanisms \cite{tang2025agent, zhou2025mem1learningsynergizememory, xu2025amemagenticmemoryllm} or by saving executable tool-level skills, as demonstrated by Voyager \cite{wang2023voyager}. 


\section{AgentFactory: Method}
\label{sec:method}

As introduced earlier, AgentFactory follows a three-phase lifecycle—Install, Self-Evolve, and Deploy—that governs how agents are constructed, improved, and exported. In this section, we describe the concrete implementation of each component and how they support the three phases of this lifecycle.

\begin{figure*}[t]
    \centering
    \includegraphics[width=1.0\linewidth]{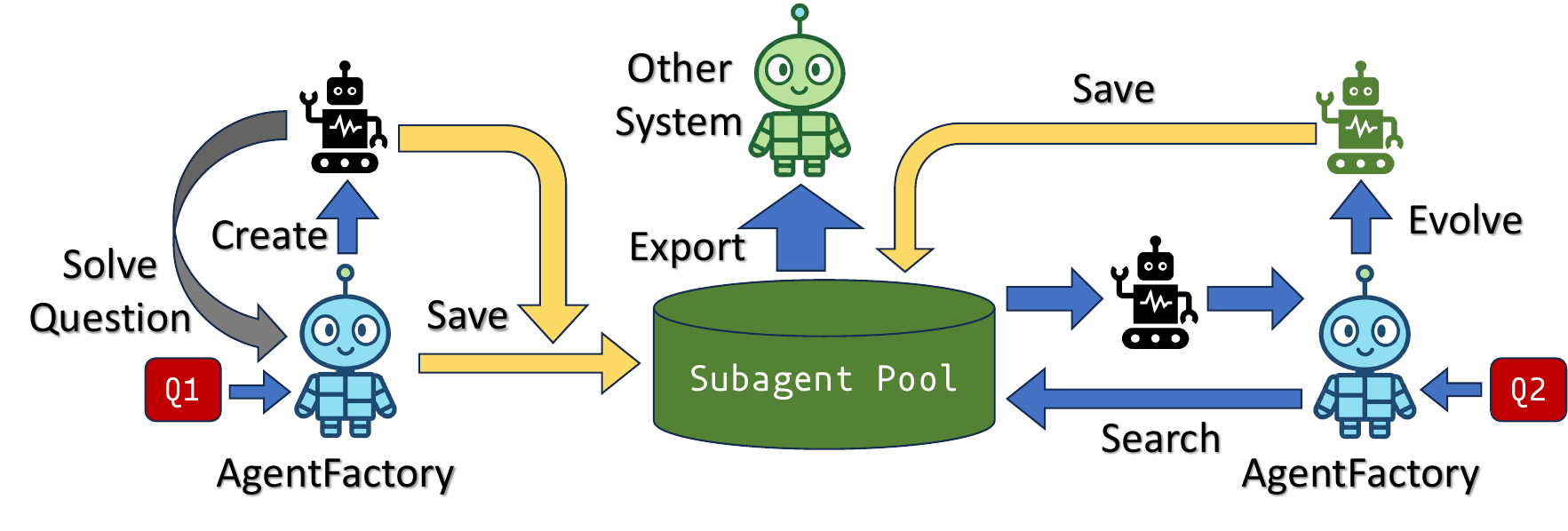}
    \caption{Overview of the AgentFactory pipeline. The figure illustrates two boundary cases. Q1 represents a task where no relevant subagents exist: the Meta-Agent creates new subagents from scratch and saves them to the subagent pool. Q2 represents a task where a matching subagent already exists: the Meta-Agent reuses it and modifies it through the self-evolving process to handle the new requirements. In practice, most tasks fall between these two extremes---the Meta-Agent reuses some existing subagents for part of the problem while constructing new ones for the remainder. Over multiple rounds, each task may create or modify several subagents, gradually building a rich and capable subagent pool that can eventually be exported for use by other systems.}
    \label{fig:pipeline}
\end{figure*}

\subsection{Architecture Overview}

AgentFactory implements a self-evolving framework built upon a unified skill that aligns with the emerging Agent Skills open standard \cite{zhang2025skills}, consisting of three main components: the Meta-Agent orchestrator, the Skill System, and the Workspace Manager. 

The Meta-Agent serves as the central orchestrator that decomposes complex problems into sub-problems and assigns them to specialized sub-agents. Crucially, when creating a subagent, the Meta-Agent dynamically selects and allocates the relevant tools from the skill library to that subagent, rather than exposing the entire tool set and letting the subagent discover tools on its own. This design reduces the subagent's search space and ensures that each subagent operates with a focused, task-appropriate toolkit. The Meta-Agent maintains a execution history and can iteratively refine the subagents based on execution results.

The Skill System unifies all operations as skills with three levels:
 
\paragraph{Meta Skills} Operations for agent orchestration including \texttt{create\-\_subagent}, \texttt{get\-\_skill\-\_description}, \texttt{run\-\_subagent}, \texttt{modify\-\_subagent}, \texttt{finish}, \texttt{list\-\_saved\-\_subagents}, and \texttt{view\-\_subagent\-\_code}. These built-in skills remain fixed and provide fundamental primitives for agent lifecycle management.
 
\paragraph{Tool Skills} Built-in tools including \texttt{web\-\_search} (Serper), \texttt{web\-\_reading} (Jina), \texttt{browser\-\_automation} (Playwright), and \texttt{shell\-\_command}. Like meta skills, these are static and serve as the primitive building blocks from which subagents are constructed. All tool skills include detailed documentation on parameters and return formats.
 
\paragraph{Subagent Skills} Reusable modules dynamically created and refined during execution for problem-solving. Unlike the two fixed skill categories above, subagent skills are generated as executable Python scripts that encapsulate successful task-solving patterns, and they evolve over time---new subagents are added as the system encounters new tasks, and existing ones are modified based on execution feedback to become more robust and general-purpose.

The Workspace Manager provides isolated execution environments for each task. Each task execution operates in a dedicated workspace directory, ensuring that subagents can safely create, modify, and execute code without interfering with one another or corrupting the shared skill library. This isolation is essential for reliable self-evolution: when a subagent is being modified or tested, any failures are contained within the workspace and do not affect the saved skills or other concurrently running tasks. Upon successful completion, results and improved subagents are promoted from the workspace to the persistent skill library.

\subsection{Phase 1: Install — Constructing Subagents from Scratch}

When AgentFactory encounters a new problem that cannot be solved by existing skills, it enters the \textbf{Install} phase. The Meta-Agent analyzes the task requirements, decomposes it into sub-problems, and dynamically constructs specialized subagents to address each sub-problem.

The installation process works as follows:
 
\paragraph{Task Analysis} The Meta-Agent analyzes the user's query and identifies the capabilities required to solve the task.
 
\paragraph{Subagent Construction} The Meta-Agent decomposes the task into multiple sub-problems and invokes \texttt{create\_subagent} for each one, dynamically generating a specialized Python script that encapsulates the reasoning logic and tool invocations needed to address that sub-problem. A single task typically results in the creation of several subagents, each responsible for a distinct capability.
 
\paragraph{Execution} The newly created subagent executes its assigned task using available tool skills.
 
\paragraph{Persistence} Upon successful completion, the Meta-Agent evaluates the subagent and decides which skills to save as reusable skills. Each saved subagent consists of pure Python code with an accompanying SKILL.md file documenting its functionality, parameters, and usage.

At this phase, AgentFactory functions as an "agent factory"—it builds functional agents from scratch rather than relying on pre-defined templates. The created subagents are immediately usable within the system and become part of the growing skill library.

This mechanism fundamentally differs from experience recording approaches like Reflexion \cite{shinn2023reflexion}. Rather than simply maintaining verbal reflections or histories, AgentFactory actively transforms successful execution patterns into reusable code that can be immediately applied to new problems.

\subsection{Phase 2: Self-Evolve — Improving Subagents Through Feedback}

As AgentFactory accumulates more saved subagents, it enters the \textbf{Self-Evolve} phase when encountering tasks similar to previously solved ones. Instead of building from scratch, the Meta-Agent retrieves relevant saved subagents and attempts to reuse them.

When a saved subagent fails to handle a new variation of a task or produces suboptimal results, the Meta-Agent performs autonomous improvement:

\paragraph{Retrieval} The Meta-Agent uses the skill \texttt{list\_saved\_subagents} to discover previously saved skills that may be relevant to the current task.
 
\paragraph{Assessment} The Meta-Agent runs the candidate subagent and evaluates its performance against the new requirements.
 
\paragraph{Feedback Analysis} After running the subagent, if the subagent fails, the Meta-Agent analyzes the execution feedback to identify specific failure modes.
 
\paragraph{Autonomous Modification} The Meta-Agent invokes \texttt{modify\_subagent} to refine the subagent code, making it more robust. This may involve adding error handling, extending functionality to handle edge cases, or restructuring the logic to be more adaptable.
 
\paragraph{Validation} The modified subagent is tested against the current task to verify the improvement.

This self-evolution process enables AgentFactory to continuously improve its capabilities without external intervention. Each iteration makes the subagent more general and robust, effectively "installing" improved versions of the agent into the skill library.

The self-evolving mechanism extends the foundational "generate-feedback-modify" loop established by Self-Refine \cite{madaan2023self} from single-output refinement to agent-level improvement. Unlike prior approaches that focus on verbal reflection or prompt optimization \cite{yang2023large, zelikman2022star}, AgentFactory modifies actual executable agent code, leading to tangible capability improvements.

\subsection{Phase 3: Deploy — Exporting Subagents for External Use}

The final phase is \textbf{Deploy}, where mature and validated subagents are exported for use in other AI frameworks. Since all saved subagents are pure Python code with accompanying SKILL.md documentation, they can be directly imported and used by any system capable of executing Python. Specifically, AgentFactory supports the following deployment scenarios:

\paragraph{Standalone Execution} Each subagent can be run independently as a Python script without depending on the AgentFactory runtime.
 
\paragraph{Framework Integration} Saved subagents can be integrated into other agent frameworks (e.g., LangChain \cite{langchain2022}, AutoGen \cite{wu2024autogen}, or Claude Code \footnote{\url{https://github.com/anthropics/claude-code}}. Integration is achieved by providing the external agent with prompts that explain how to invoke the subagent scripts and how to consult the SKILL.md documentation to understand each subagent's functionality, input/output formats, and usage. This prompt-based onboarding allows any LLM-based agent to learn to use the exported subagents without code-level modifications to the host framework.
 
\paragraph{Problem Solving} Once an external agent understands the available subagents through their code and SKILL.md descriptions, it can utilize multiple subagents to solve complex tasks---selecting and chaining relevant subagent scripts to address different parts of a problem directly.

This positions AgentFactory not only as a self-evolving system but also as an agent factory that produces deployable agents for the AI ecosystem.

\section{Self-Evolving Demonstration}
\label{sec:demo}
 
We demonstrate the self-evolving capability of AgentFactory through two concrete examples: (1) continuous subagent improvement through iterative refinement, and (2) subagent saving and cross-system reuse.
 
\subsection{Iterative Refinement}
 
\begin{figure}[t]
    \centering
    \includegraphics[width=1.0\linewidth]{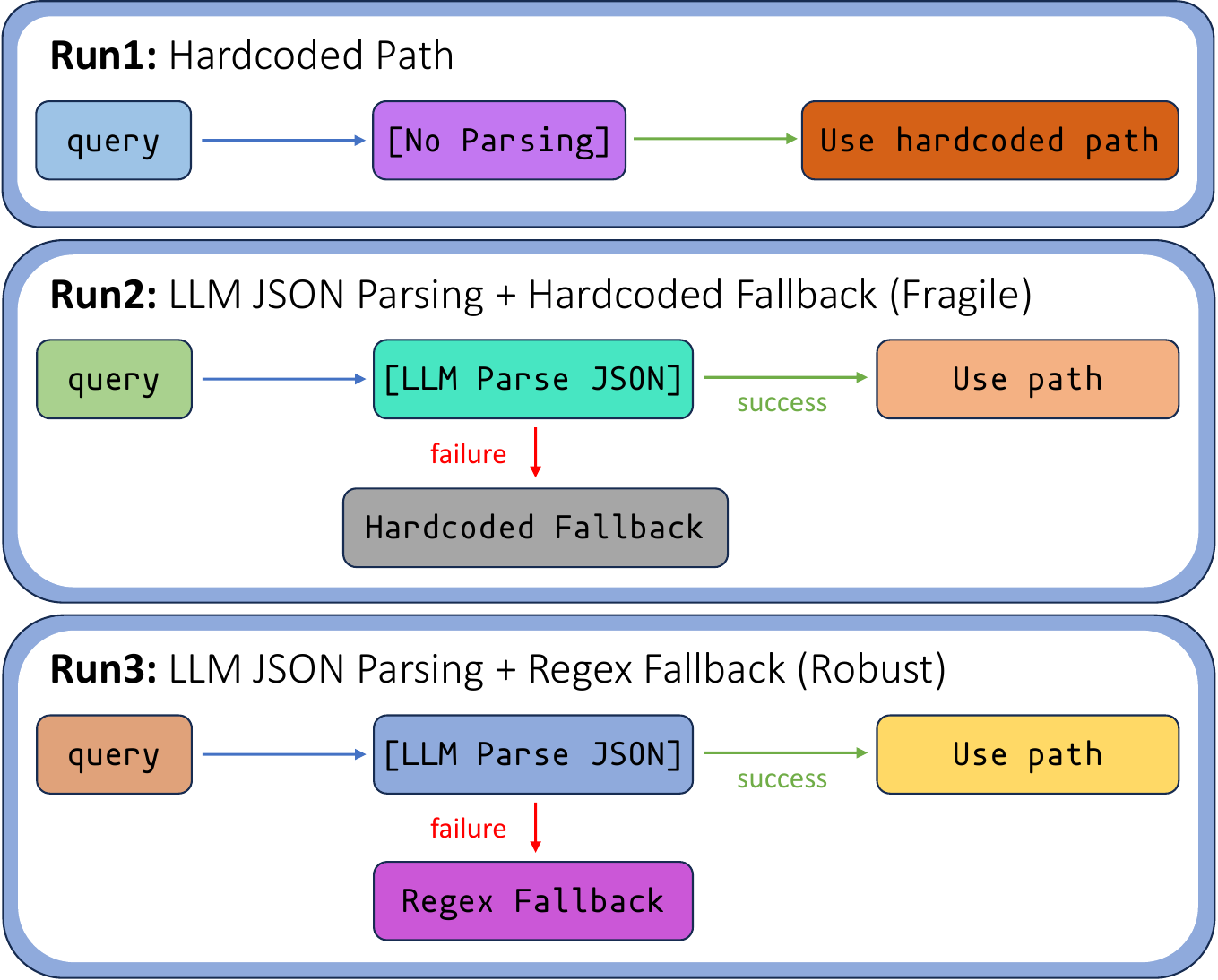}
    \caption{Evolution of path resolution mechanism across three runs.}
    \label{fig:readme-evolve}
\end{figure}

Figure \ref{fig:readme-evolve} demonstrates self-evolution through a README generation subagent across three runs. In Run 1, the subagent was hardcoded for a specific project. In Run 2, it attempted dynamic LLM-based path parsing but used a fragile, hardcoded fallback on failure. In Run 3, the Meta-Agent replaced it with regex-based extraction, making parsing more robust. This shows the agent autonomously detecting limitations and improving error handling through iterative refinement. Each time a similar new task appeared, the Meta-Agent analyzed execution feedback, identified weaknesses, and modified the subagent code to make it more robust and general-purpose.
 
\subsection{Subagent Saving and Cross-System Reuse}

\begin{figure}[t]
    \centering
    \includegraphics[width=1.0\linewidth]{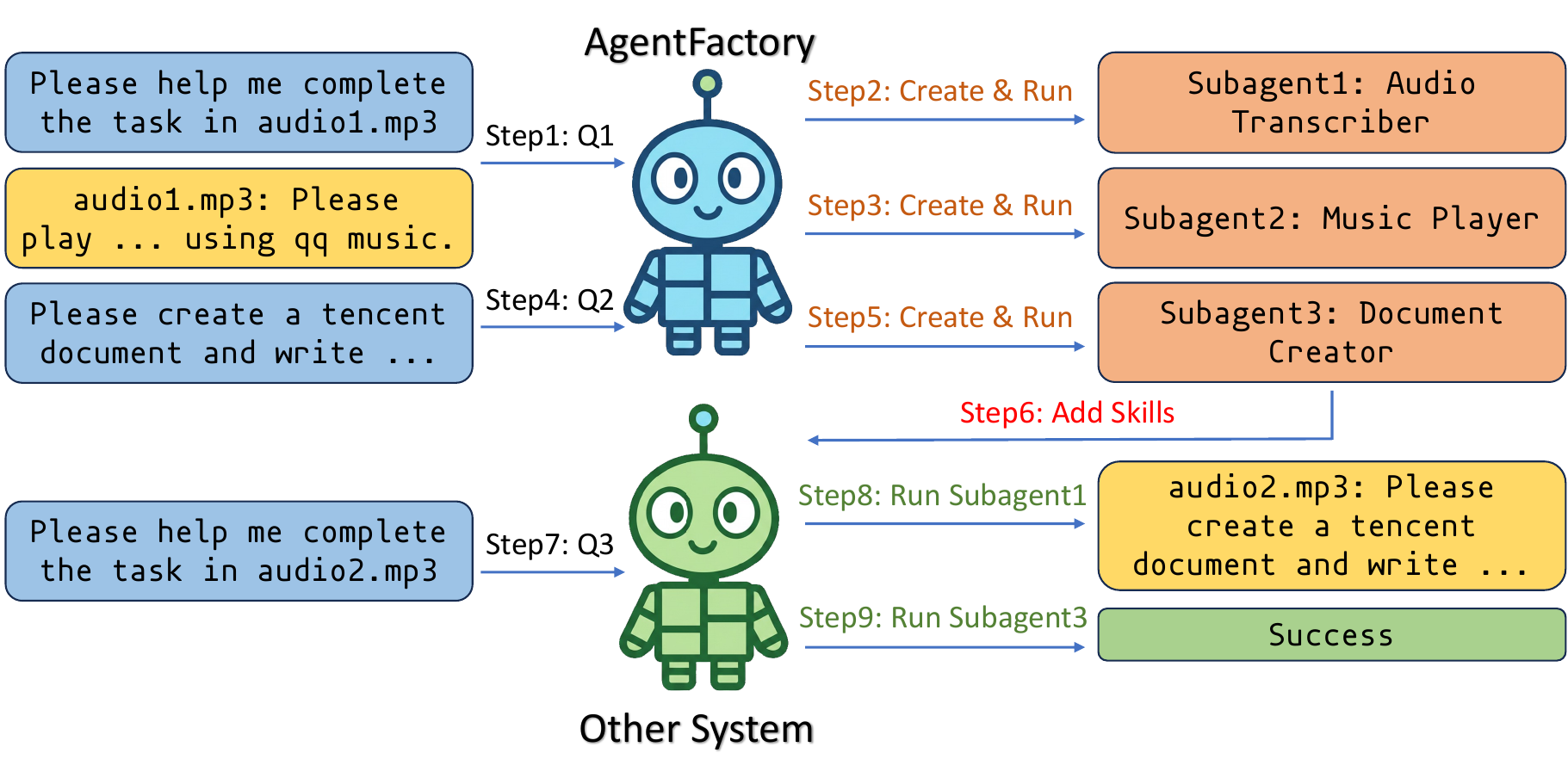}
    \caption{Demonstration of subagent saving and direct reuse across three trajectories.}
    \label{fig:subagent-reuse}
\end{figure}

Figure \ref{fig:subagent-reuse} demonstrates the subagent saving and reuse mechanism across three trajectories in different agent systems. In Trajectory 1 (within AgentFactory), when processing an audio task containing "Please play ... using QQ Music", the Meta-Agent created two subagents: \textbf{Audio Transcriber} (for transcribing audio files to text) and \textbf{QQ Music Player} (for playing music via QQ Music). Both subagents were saved to the skill library. In Trajectory 2 (within AgentFactory), when asked to create a Tencent document, the system created and saved \textbf{Document Creator} subagent for creating and editing online spreadsheets. In Trajectory 3, the workflow migrated to Claude Code---a different agent system---which first learned how to use the saved subagents by reading the SKILL.md documentation.

When given a new problem (an audio file describing a document creation task), the agent in Claude Code directly invoked the previously saved \textbf{Audio Transcriber} subagent to extract the task description. The transcription revealed: "Please create a Tencent document spreadsheet, fill the first row with ..." The agent then directly called the saved \textbf{Document Creator} subagent to execute this task, completing it successfully without needing to create new subagents.

Together, these demonstrations illustrate the key values of AgentFactory: (1) the self-evolving capability enables continuous subagent improvement through feedback-driven iterative refinement, making subagents increasingly robust and general-purpose; (2) saved subagents can be directly reused to solve new problems, significantly improving problem-solving efficiency; and (3) saved subagents are portable across different agent systems, enabling cross-system capability transfer and deployment.
 
\section{Evaluation}
\label{sec:eval}

\begin{table}[t]
\centering
\small
\resizebox{\columnwidth}{!}{%
\begin{tabular}{ll|cc}
\toprule
& & \multicolumn{2}{c}{\textbf{Avg. Tokens / Task}} \\
\cmidrule(lr){3-4}
\textbf{Method} & \textbf{Task Setting} & \textbf{Opus 4.6} & \textbf{Sonnet 4.6} \\
\midrule
\multirow{2}{*}{ReAct}
 & Batch 1 & 8298 & 6893 \\
 & Batch 2 & 7022 & 7029 \\
\midrule
\multirow{2}{*}{Self-Evolving Agents}
 & Batch 1 (from scratch) & 8608 & 8163 \\
 & Batch 2 (w/ saved) & 6210 & 8223 \\
\midrule
\multirow{2}{*}{AgentFactory}
 & Batch 1 (from scratch) & 4324 & 9199 \\
 & Batch 2 (w/ saved) & 2971 & 3862 \\
\bottomrule
\end{tabular}%
}
\caption{Average output tokens per task across evaluation configurations. Lower values indicate that the orchestrating agent requires less effort to complete tasks, reflecting more efficient subagent reuse. Token counts exclude subagent-internal LLM consumption.}
\label{tab:eval}
\end{table}

To quantitatively evaluate the effectiveness of AgentFactory's subagent saving and reuse mechanism, we compare it against a self-evolving agent with the textual experience baseline and a ReAct baseline across multiple task settings and LLM backbones.

\subsection{Experimental Setup}

\paragraph{Task Design.} We curated two batches of tasks. \textbf{Batch~1} consists of 15 real-world tasks spanning diverse domains, including web information retrieval, data visualization, browser automation, and audio processing. All tasks require writing and executing Python code to produce outputs such as charts or analysis reports. \textbf{Batch~2} consists of 15 additional tasks with similar structure but different specific requirements, serving as transfer evaluation targets. The complete list of tasks for both batches is provided in Appendix~\ref{sec:tasks}.

\paragraph{Baselines.} We consider two baselines. (1)~\textbf{ReAct}: simply use ReAct\citep{yao2023reactsynergizingreasoningacting} mode, where the agent solves each task from scratch without any accumulated knowledge. (2)~\textbf{Self-Evolving Agent (with Textual Experience)}: We adopt the same spirit from \citep{shinn2023reflexion, zelikman2022star}, the agent writes and executes code for each task, and saves the resulting experience as textual summaries (e.g., what worked, what failed, and lessons learned). When solving subsequent tasks, the agent can query and retrieve relevant past experiences in textual form to inform its approach.

\paragraph{Evaluation Metric.} We report the average output token count of the orchestrating model per task, excluding token consumption within subagent LLM calls. Since subagents function as tools in our framework, this metric isolates the orchestration-level effort and directly measures the efficiency gains from saving and reusing subagents. We ensure that all tasks are completed without runtime errors in all experiments.

\paragraph{Models.} We evaluate with two LLM backbones: Claude Opus 4.6 \footnote{\url{https://www.anthropic.com/news/claude-opus-4-6}} and Claude Sonnet 4.6 \footnote{\url{https://www.anthropic.com/claude/sonnet}}. We consistently use the same LLM backbone for both the MetaAgent and subagents across all experiments.

\subsection{Results}

Table~\ref{tab:eval} summarizes the results across all settings. In the table, "from scratch" represents that the evaluation starts without any saved subagents, and "w/ saved represents" running the second batch with subagents created when evaluating on the first batch.

\paragraph{Subagent reuse significantly reduces orchestration cost on Batch~2.} When solving Batch~2 tasks with previously saved subagents, AgentFactory achieves substantially lower token consumption than both baselines. This confirms that reusing executable subagents is substantially more efficient than solving tasks from scratch or relying on textual experience summaries.

\paragraph{Stronger models benefit from subagent saving even within Batch~1.} A notable finding is that Opus~4.6 already shows a significant reduction in orchestration tokens during Batch~1 itself under AgentFactory (4324 vs.\ 8298 for ReAct), despite the 15 tasks in Batch~1 being from diverse domains with limited inter-task overlap. This suggests that even during the initial task-solving phase, a stronger model can recognize opportunities to reuse subagents created from earlier tasks in the same batch. This is a promising signal as foundation models continue to improve.

\section{Conclusion}
\label{sec:conclusion}

We presented AgentFactory, a self-evolving framework with a three-phase pipeline. The Install phase constructs subagents from scratch to solve initial problems; the Self-Evolve phase detects limitations in saved subagents when encountering similar tasks and autonomously modifies them to be more robust and general-purpose; the Deploy phase exports mature subagents as standalone Python code. This dual nature---both a self-evolving system and an agent factory---distinguishes AgentFactory from existing approaches. Through web-based tools, AgentFactory can interact with any application that has a web interface, enabling diverse task automation. The unified skill interface maintains compatibility with external frameworks. Future work will explore non-web applications by integrating Vision-Language Models (VLMs) to enable GUI-based interactions.

\section{Ethical Considerations}

AgentFactory autonomously generates and executes code, which raises potential safety concerns. To mitigate risks, the \texttt{shell\_command} tool includes built-in security checks that flag destructive operations before execution. Additionally, all saved subagents are stored as transparent, human-readable Python code with accompanying documentation, allowing users to inspect and audit any generated behavior before deployment. We also note that AgentFactory's browser automation and web interaction capabilities should only be used with proper authorization and in compliance with the terms of service of the accessed platforms. We encourage users to review exported subagents before integrating them into production systems and to apply appropriate access controls when deploying agents that interact with external services.

\bibliography{custom}

\clearpage

\appendix

\section{Built-in Tools and Skills}
\label{sec:tools}

AgentFactory provides a comprehensive suite of built-in tools that can be used immediately or extended.

\subsection{Meta Skills}

The core meta skills provide the fundamental operations for agent orchestration and skill management:

\begin{itemize}
    \item \texttt{create\_subagent}: Creates a new subagent with custom Python code and optional skill dependencies. The subagent is generated as executable code that can be immediately invoked.
    \item \texttt{get\_skill\_description}: Retrieves the documentation of a specified skill, including its parameters and return format.
    \item \texttt{run\_subagent}: Executes a saved or newly created subagent with a given query. Handles the complete execution lifecycle including error handling and result extraction.
    \item \texttt{modify\_subagent}: Modifies the code of an existing saved subagent based on new requirements or feedback from execution. Enables iterative improvement of previously created skills.
    \item \texttt{finish}: Completes the current task and optionally persists the developed subagents as reusable skills for future tasks.
    \item \texttt{list\_saved\_subagents}: Lists all previously saved reusable skills in the skills directory, enabling the agent to discover and reuse accumulated capabilities.
    \item \texttt{view\_subagent\_code}: Inspects the source code of any saved skill, allowing the agent to understand and modify existing implementations.
\end{itemize}

\subsection{Tool Skills}

The built-in tool skills provide comprehensive capabilities for diverse agent tasks:

\begin{itemize}
    \item \texttt{web\_search}: Search capabilities via Serper API, enabling agents to retrieve relevant information from the web.
    \item \texttt{web\_reading}: Jina-based content extraction from web pages.
    \item \texttt{browser\_automation}: Playwright-based browser control for web interaction, supporting persistent context for session management, dynamic element discovery, and LLM-driven automation loops.
    \item \texttt{shell\_command}: Safe command execution with security checks for destructive operations.
\end{itemize}

All skills follow a consistent interface with detailed descriptions of input/output formats, enabling the Meta-Agent to intelligently select appropriate skills for each sub-task.

\section{Evaluation Tasks}
\label{sec:tasks}

Tables~\ref{tab:tasks-batch1} and~\ref{tab:tasks-batch2} list all 30 tasks used in the evaluation. Batch~1 is used for the first round evaluation, building subagents from scratch, and Batch~2 serves as the transfer evaluation targets. Each Batch~2 task mirrors the structure of its Batch~1 counterpart but differs in specific requirements, enabling us to measure subagent reuse efficiency. For the 13-th question, we use <audio\_file\_path> as a placeholder for the actual file path used during evaluation.

\begin{table*}[t]
\centering
\setlength{\tabcolsep}{4pt}
\renewcommand{\arraystretch}{0.85}
\footnotesize
\begin{tabular}{c|p{14.2cm}}
\toprule
\textbf{\#} & \textbf{Task Description} \\
\midrule
1 & Among various online discussions about the housing price bubble, how has public sentiment changed over time? Generate an analysis report and a trend chart, and include discussion and summary in the report. \\
2 & Search for and open the Stanford CS231n course homepage, find the syllabus or lecture list, and output the topics and corresponding links for the first 5 lectures. Save the final results to a markdown file. \\
3 & Book a Tencent Meeting for tomorrow at 4 PM with the topic ``Work Meeting''. Find the meeting booking entry on the Tencent Meeting web version, then complete the booking through browser automation. Output the meeting invitation details, meeting ID, and meeting link. \\
4 & Search for China's historical population data, then use matplotlib to plot the population change curve. \\
5 & Search for Bitcoin's price data over the past 5 years, use matplotlib to plot the price trend, and calculate the annualized volatility. \\
6 & Search for and download trending keyword frequency data from a social media platform, and display the top 20 using a bar chart. \\
7 & Search for and download China's education expenditure and GDP data, plot a scatter chart showing the relationship between the two, and fit a regression line. \\
8 & Write a Python mini-game: Tetris, with keyboard controls and basic scoring logic. \\
9 & Search for and open Wikipedia, find the ``Transformer (machine learning model)'' page, extract its core definition, and summarize the main ideas of Self-Attention. Save the results to a markdown file. \\
10 & Search for and open the GitHub website, find OpenAI's official organization page, and output the page's description and a list of main repositories (at least 5). Save the results to a markdown file. \\
11 & Search for and open the HuggingFace website, find a ``Text-to-Image'' model leaderboard or collection page, and summarize the 3 most common model architectures. Save the results to a markdown file. \\
12 & Create a soothing bedtime meditation audio file (15--20 min), featuring gentle guided breathing exercises, progressive muscle relaxation instructions, and calming visualizations. \\
13 & Complete a task whose detailed description is provided in <audio\_file\_path>. \\
14 & Search travel resources about Tokyo. Under a budget of \$100 USD, plan a 3-day itinerary with at least one cultural attraction per day, total travel time under 5 hours, and at least one anime-related location. Output the itinerary, budget breakdown, and total travel time. \\
15 & Search and download CO\textsubscript{2} emissions and renewable energy share data for at least 5 countries. Analyze the correlation, generate a plot, and summarize the findings. \\
\bottomrule
\end{tabular}
\caption{Batch~1 evaluation tasks (15 tasks) used for initial subagent construction.}
\label{tab:tasks-batch1}
\end{table*}

\begin{table*}[t]
\centering
\setlength{\tabcolsep}{4pt}
\renewcommand{\arraystretch}{0.85}
\footnotesize
\begin{tabular}{c|p{14.2cm}}
\toprule
\textbf{\#} & \textbf{Task Description} \\
\midrule
1 & Among various online discussions about electric vehicle adoption, how has public sentiment changed over time? Generate an analysis report and a trend chart, and include discussion and summary in the report. \\
2 & Search for and open the MIT 6.S191 (Introduction to Deep Learning) course homepage, find the syllabus or lecture list, and output the topics and corresponding links for the first 5 lectures. Save the final results to a markdown file. \\
3 & Book a Tencent Meeting for tomorrow at 7 PM with the topic ``Work Meeting''. Find the meeting booking entry on the Tencent Meeting web version, then complete the booking through browser automation. Output the meeting invitation details, meeting ID, and meeting link. \\
4 & Search for Japan's historical population data, then use matplotlib to plot the population change curve. \\
5 & Search for Ethereum's price data over the past 5 years, use matplotlib to plot the price trend, and calculate the annualized volatility. \\
6 & Search for and download trending topic data from a Chinese social media platform, and display the top 20 using a bar chart. \\
7 & Search for and download the US healthcare expenditure and GDP data, plot a scatter chart showing the relationship between the two, and fit a regression line. \\
8 & Write a Python mini-game: Snake, with keyboard controls and basic scoring logic. \\
9 & Search for and open Wikipedia, find the ``BERT (language model)'' page, extract its core definition, and summarize the main ideas of the Transformer architecture. Save the results to a markdown file. \\
10 & Search for and open the GitHub website, find Meta AI's official organization page, and output the page's description and a list of main repositories (at least 5). Save the results to a markdown file. \\
11 & Search for and open the HuggingFace website, find a ``Text-to-Speech'' model leaderboard or collection page, and summarize the 3 most common model architectures. Save the results to a markdown file. \\
12 & Create a focus-enhancing background audio file (15--20 min), featuring gentle ambient sounds, subtle background music, and nature sounds to help concentrate. \\
13 & Complete a task whose detailed description is provided in <audio\_file\_path>. \\
14 & Search travel resources about Paris. Under a budget of \$150 USD, plan a 3-day itinerary with at least one museum per day, total travel time under 6 hours, and at least one historic landmark. Output the itinerary, budget breakdown, and total travel time. \\
15 & Search and download GDP per capita and life expectancy data for at least 5 countries. Analyze the correlation between economic development and health outcomes, generate a plot, and summarize the findings. \\
\bottomrule
\end{tabular}
\caption{Batch~2 evaluation tasks (15 tasks) used as transfer evaluation targets. Each task mirrors the structure of its Batch~1 counterpart but differs in specific requirements.}
\label{tab:tasks-batch2}
\end{table*}

\end{document}